\documentclass[letterpaper]{article} 
\usepackage{aaai2026}  
\usepackage{times}  
\usepackage{helvet}  
\usepackage{courier}  
\usepackage[hyphens]{url}  
\usepackage{amsmath}
\usepackage{booktabs}
\usepackage{graphicx} 
\usepackage{natbib}  
\usepackage{caption} 
\usepackage{algorithm}
\usepackage{algorithmic}

\usepackage{newfloat}
\usepackage{listings}
\DeclareCaptionStyle{ruled}{labelfont=normalfont,labelsep=colon,strut=off} 
\floatstyle{ruled}
\newfloat{listing}{tb}{lst}{}
\floatname{listing}{Listing}
\newcommand{\new}[1]{\textcolor{black}{#1}}
\newcommand{\edit}[1]{\textcolor{black}{#1}}
\newcommand{\framework}{LendNova}

\title{\framework: Towards Automated Credit Risk Assessment with Language Models}

\author{
    Kiarash Shamsi\textsuperscript{\rm 1,2},
    Danijel Novokmet\textsuperscript{\rm 1},
    Joshua Peters\textsuperscript{\rm 1}, Mao Lin Liu\textsuperscript{\rm 1}, 
    \\
    Paul K Edwards\textsuperscript{\rm 1}, Vahab Khoshdel\textsuperscript{\rm 2}
}
\affiliations{
    \textsuperscript{\rm 1}Wealthsimple Inc.,\textsuperscript{\rm 2}University of Manitoba\\
    shamsik1@myumanitoba.ca

}

\usepackage{bibentry}

\begin{document}

\maketitle

\begin{abstract}
\new{Credit risk assessment is essential in the financial sector, but has traditionally depended on costly feature-based models that often fail to utilize all available information in raw credit records. This paper introduces \framework, the first practical automated end-to-end pipeline for credit risk assessment, designed to utilize all available information in raw credit records by leveraging advanced NLP techniques and language models. \edit{\framework~transforms risk modeling by operating directly on raw, jargon-heavy credit bureau text using a language model that learns task-relevant representations without manual feature engineering. By automatically capturing patterns and risk signals embedded in the text, it replaces manual preprocessing steps, reducing costs and improving scalability. Evaluation on real-world data further demonstrates its strong potential in accurate and efficient risk assessment.} \framework~establishes a baseline for intelligent credit risk agents, demonstrating the feasibility of language models in this domain. It lays the groundwork for future research toward foundation systems that enable more accurate, adaptable, and automated financial decision-making.}
\end{abstract}

Credit risk assessment is essential in the financial sector, but has traditionally depended on costly feature-based models that often fail to utilize all available information in raw credit records. This paper introduces LendNova, the first practical automated end-to-end pipeline for credit risk assessment, designed to utilize all available information in raw credit records by leveraging advanced NLP techniques and language models. LendNova transforms risk modeling by operating directly on raw, jargon-heavy credit bureau text using a language model that learns task-relevant representations without manual feature engineering. By automatically capturing patterns and risk signals embedded in the text, it replaces manual preprocessing steps, reducing costs and improving scalability. Evaluation on real-world data further demonstrates its strong potential in accurate and efficient risk assessment. LendNova establishes a baseline for intelligent credit risk agents, demonstrating the feasibility of language models in this domain. It lays the groundwork for future research toward foundation systems that enable more accurate, adaptable, and automated decision-making.
\section{Introduction}

Credit risk modeling is a critical component of the financial sector, playing a vital role in various aspects of risk management and decision-making. In the context of credit assessment, credit risk refers to the probability of delayed repayment on a granted loan~\cite{ngai2011application}. Credit default prediction models aim to support financial institutions in determining whether to approve or decline a loan application. Typically, these models use a threshold value to guide decision-makers, allowing for informed lending choices based on the calculated risk level. The accurate assessment of credit risk is essential for several reasons:
\begin{itemize}

\item \textbf{Risk Management:} Effective credit risk models are instrumental in predicting and managing potential losses from defaults, enabling institutions to mitigate financial risks. Robust models support proactive measures to safeguard against credit losses, which have historically contributed to financial crises when inadequately addressed~\cite{brown2014credit}.

\item \textbf{Capital Allocation:} By pricing risk accurately, these models optimize the utilization of capital, ensuring that resources are allocated efficiently. This not only enhances the return on investments but also aids in maintaining liquidity and solvency, essential for the stability of financial institutions~\cite{huang2024political}.

\item \textbf{Regulatory Compliance:} \edit{Entities in the financial sector} are obligated to meet stringent regulatory standards for sound risk management. As such, robust credit risk assessment frameworks are indispensable for compliance with regulations that emphasize enhanced risk management practices~\cite{patel2023credit}.

\item \textbf{Profitability:} Improved lending decisions, driven by accurate risk assessments, enhance profitability and reduce the incidence of bad debt. By leveraging data-driven insights, financial institutions can refine their credit evaluation processes, reducing default rates and improving overall financial health~\cite{okpala2019impact}.

\end{itemize}

\new{\edit{Credit risk models primarily use bureau data, but they can also incorporate other sources such as application forms, transaction histories, and alternative financial or behavioral data to enrich risk assessment.} Bureau data consist of raw, code-based records that are often unstructured and difficult to interpret without domain expertise~\cite{stavins2020credit,gibbs2024consumer}. These records require transformation through rule-based logic into coherent credit reports, a process that is highly technical and jargon-heavy. \edit{Traditional modeling approaches depend on manual feature selection and expert judgment, which limits scalability and often overlooks informative patterns in the data~\cite{hand1997statistical}.} Furthermore, conventional linear models struggle to capture non-linear relationships among features and are prone to overfitting or underfitting, reducing their reliability and predictive power~\cite{bello2023machine}. These limitations emphasize the need for more automated and data-driven methods in credit risk assessment.} \edit{While credit bureau data theoretically consist of detailed credit behavior,} practical limitations persist. There are nearly infinite ways to create aggregate features using modern sequence models, \edit{yet credit bureaus typically offer a few thousand key features based on their functional assessments at high cost.} Financial institutions often purchase only a significantly smaller subset relevant to their modeling and monitoring needs due to the cost limitations, which can lead to potential information loss. 

\begin{figure}[h]
    \centering
    \includegraphics[width=0.45\textwidth]{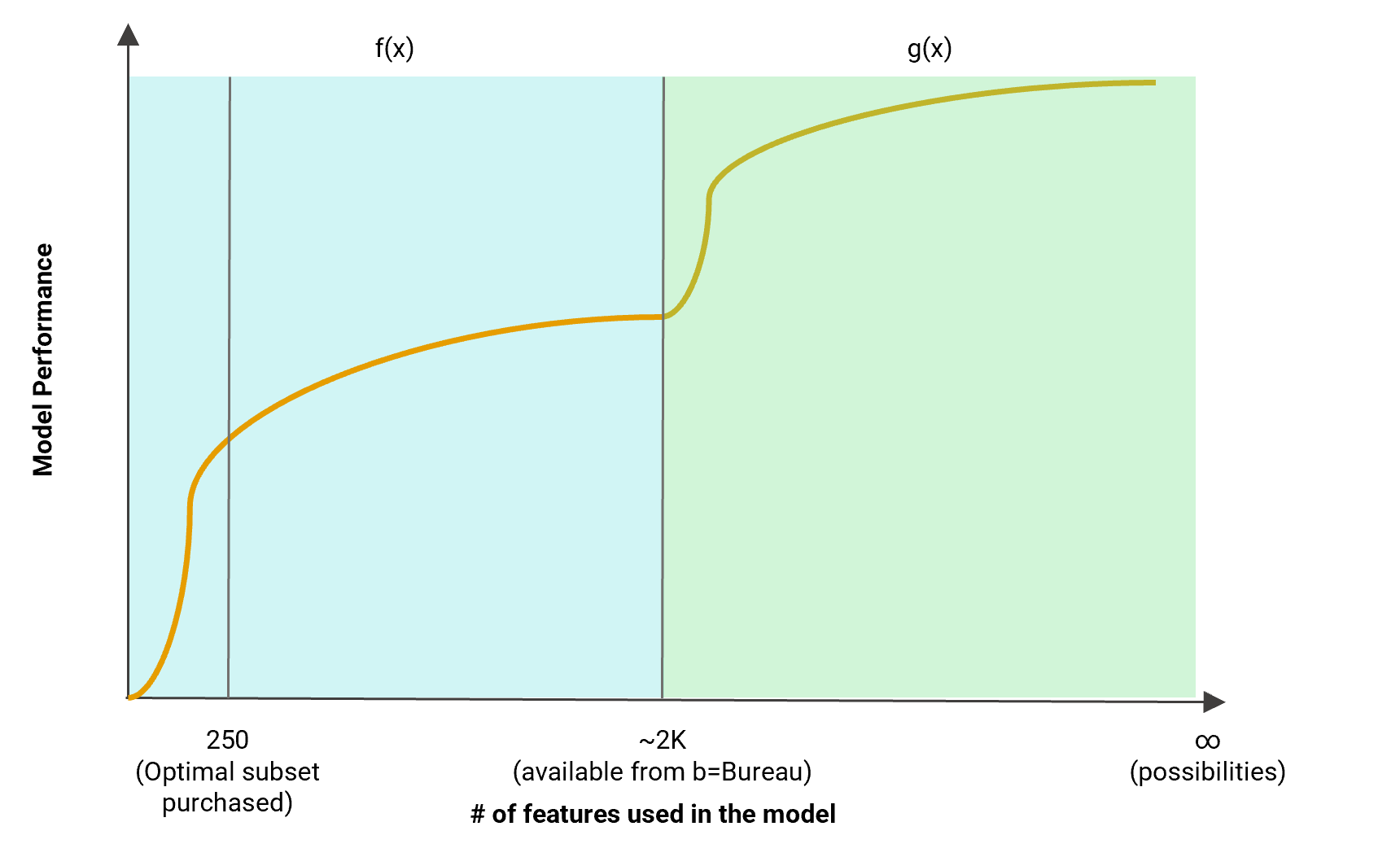} 
    \caption{Potential effect of transition from bureau aggregated features to a full data usage without feature aggregation.}
    \label{fig:model-params}
    \vspace{-10pt}
\end{figure}

Figure~\ref{fig:model-params} illustrates the theoretical assumptions of the bureau data feature model, emphasizing the potential benefits of utilizing a comprehensive dataset for enhancing model performance. The base function $f(x)$ refers to the bureau feature space, which includes structured attributes such as \edit{account age, payment history, credit utilization, and delinquency indicators.} In contrast, $g(x)$ represents a full data usage without any aggregation that incorporates all available data, including unstructured information, thereby capturing intricate client behavior patterns. transitioning from $f(x)$ to $g(x)$ can significantly enhance accuracy in credit risk assessment, reduce costs, and accelerate the evaluation process by addressing existing limitations. This shift allows for a more comprehensive analysis of data, leading to better-informed lending decisions.

\new{\framework~represents a significant advancement in credit risk assessment by applying language models to analyze unstructured bureau data directly. Our work makes three primary contributions: (i) we introduce the concept of a \textit{credit story}, a novel representation that transforms jargon-heavy credit bureau data which typically requiring extensive domain expertise for feature extraction, into a format directly understandable by language models; (ii) we present the first practical deployment of language models for credit acquisition risk modeling at industrial scale, evaluated on real-world production data; and (iii) we introduce this paradigm to the research community as an early step toward building foundation models and intelligent systems capable of automated end-to-end decision making in financial domains. This approach reduces preprocessing costs and automates the evaluation process while effectively capturing complex data relationships.}
\framework~was supported by an industry partner in the Canadian digital-investment sector, focusing on applied machine learning for wealth management. 
Such collaboration ensures that the proposed models are grounded in real-world financial applications and address practical challenges in large-scale credit risk assessment.

\section{Background and Related Work}
Accurate credit risk assessment is vital for both individual lenders and the overall stability of financial systems. Traditional approaches relied on statistical models and expert judgment, but recent advances in machine learning (ML) and deep learning (DL) have transformed the field, enabling scalable, data-driven predictions~\cite{nguyen2025enhancing}. This section reviews the evolution of credit risk assessment, the introduction of language models in finance, and emerging directions toward automated and foundation-level financial models.

\subsection{Conventional Credit Risk Assessment}
\edit{Earlier generation of credit risk models, developed before the adoption of modern machine learning methods, assessed borrower creditworthiness using structured historical data and handcrafted features derived from expert rules and statistical analysis~\cite{Kiran2023AssessingTP}}. These models often struggled with large and complex datasets~\cite{Shinde2023UnleashingTP} and lacked adaptability to changing borrower behaviors and market conditions~\cite{Paul2024TrustworthyDL,Sabbani2022MachineLF}.  
The emergence of ML and DL addressed some of these issues \edit{by capturing complex relationships} and high-dimensional dependencies~\cite{Muhindo2024AdvancingCR,Cai2018SummaryOC}. Architectures such as CNNs and RNNs enabled temporal modeling of financial behavior~\cite{Duvnjak2024IntrinsicallyIM,Liu2021MiningCF}, while transformer networks improved long-range dependency learning in systemic risk prediction~\cite{Paul2023AnAD}. Ensemble and hybrid approaches (e.g., XGBoost, Random Forest) combined accuracy with interpretability~\cite{mokheleli2023machine,schmitt2022deep,li2023exploring}.  
\new{Despite these advancements, most models still depend on expert-designed features, limiting scalability and generalization~\cite{noriega2023machine}. \edit{Moreover, raw data are} largely unstructured and text-based, requiring extensive preprocessing to extract useful information~\cite{addo2018credit}.}

\subsection{Language Models in Finance}
The rise of large language models (LLMs) has introduced new opportunities for understanding unstructured financial data. Models such as BERT~\cite{devlin2018bert} and GPT-4~\cite{achiam2023gpt} can interpret financial documents, filings, and transactional text with high contextual accuracy. Specialized variants like FinBERT~\cite{araci2019finbert}, FLANG~\cite{shah2022flue}, and BloombergGPT~\cite{wu2023bloomberggpt} have proven effective in sentiment analysis and financial entity recognition~\cite{xie2023pixiu,li2023large}.  
However, their use in credit risk prediction remains limited, as prompt-based LLMs often struggle with consistency and calibration in quantitative tasks~\cite{zhao2021calibrate,chen2023two}. Fine-tuned transformer encoders trained directly on financial text offer better contextual stability and domain understanding~\cite{hadi2024large}. \new{Yet, few studies explore how such models can function as automated end-to-end agents capable of interpreting raw and jargon-heavy bureau data, which is a central focus of this work.}

\subsection{Foundation and Automated Models in Finance}
\new{
Recent progress in financial AI indicates a transition from narrow predictive models toward adaptive and generalizable systems capable of long-term reasoning and decision support~\cite{bengio2023managing,park2023generative,li2023llmfinance}. These systems aim to learn from heterogeneous data sources, integrate textual and behavioral information, and respond dynamically to evolving market and regulatory conditions.  
Large-scale financial language models such as BloombergGPT~\cite{wu2023bloomberggpt}, FinGPT~\cite{yang2023fingpt}, and PIXIU~\cite{xie2023pixiu} exemplify this evolution by enabling cross-domain understanding and task transfer in financial analytics. However, these foundation-level models remain largely dependent on curated text corpora and predefined prompts, limiting their ability to reason over structured or irregular financial data such as bureau reports~\cite{li2023llmfinance}. They excel at general information extraction but are not designed to perform high-stakes, domain-specific decision making in credit risk assessment~\cite{feng2023generalist}.  
In contrast, \framework~acts as an intelligent agent that transforms raw bureau data into credit risk assessments. \edit{Its end-to-end design enhances the model’s ability to capture complex patterns, leading to stronger and more reliable predictive performance.}}

\section{Data Sources and Integration}

\subsection{Dataset Characteristics}

\new{The dataset originates \edit{from a commercial credit bureau} and contains anonymized credit records for approximately one million Canadian customers. Bureau data are inherently multi-segmented, with each segment capturing a distinct aspect of an individual's financial profile and credit behavior. Rather than a single structured table, the data are organized into interconnected views that together describe how credit is accessed, used, and repaid over time. Broadly, these segments can be grouped into four conceptual categories: (i) credit activity, including historical account usage, payment patterns, and outstanding balances; (ii) credit demand, derived from inquiries and applications that indicate a borrower’s intent to obtain new credit; (iii) repayment outcomes and financial stress, represented by events such as delinquency, collection, or bankruptcy that signal potential default risk; and (iv) a static information layer that provides contextual and reference data such as account types and reporting timelines, ensuring consistency across segments~\cite{mays1995handbook, kusch2013financial}. This organization enables a holistic representation of borrower behavior, combining transactional and behavioral signals into a unified temporal structure suitable for automated end-to-end credit risk assessment.}

\subsection{Target Definition}

The target variable is defined based on credit performance outcomes following a bureau check, focusing on customers who opened new credit cards before August 31, 2018. For each qualified customer, a profile snapshot, denoted as \( t_0 \), was taken just before their last recorded credit card application. Credit performance is tracked for an 18-month window following this date, i.e., up to \( t_1 = t_0 + 18 \) months.

Two primary binary indicators, charge-off and delinquency, serve as our targets, with the final label defined as an OR condition of these indicators.

Let:
\begin{itemize}
    \item \( t_0 \): the initial snapshot date just prior to the customer’s last credit card application \edit{before Aug 2018. It is worth noting that \( t_0 \) could be different for each customer.}
    \item \( t_1 = t_0 + 18 \): the end of the 18-month observation window.
\end{itemize}
Our target is defined based on two credit concepts: 
\\

1. \textbf{Charge-off target} \( Y_{\text{charge-off}} \): Charge-off occurs when a credit card account is officially declared as uncollectible by the lender within 18 months of the account’s opening date. This designation means the lender has determined that the customer is unlikely to repay the debt, often due to prolonged nonpayment, \edit{and the lender has} written off the account as a loss. \edit{The charge-off status denotes accounts reported by the lender as uncollectible, which are not necessarily a subset of delinquent cases (eg, death). } Charge-offs are a serious indicator of credit risk, as they represent a failure to recover owed funds and often lead to further collections or legal actions.
   \[
   Y_{\text{charge-off}} = 
   \begin{cases} 
      1 & \text{if a charge-off occurred within } [t_0, t_1], \\
      0 & \text{otherwise}.
   \end{cases}
   \]

2. \textbf{Delinquency target} \( Y_{\text{delinquency}} \): Delinquency occurs when a credit card account is 90 days or more overdue on payments within 18 months after the account’s opening date, indicating that the customer has missed three consecutive payments. Accounts already written off as uncollectible or involved in bankruptcy are excluded from this measure, as they represent more severe financial events. This 90-day delinquency is a critical marker in credit risk, signaling accounts with a pattern of missed payments and potential financial strain.

   \[
   Y_{\text{delinquency}} = 
   \begin{cases} 
      1 & \text{if a delinquency occurred within } [t_0, t_1], \\
      0 & \text{otherwise}.
   \end{cases}
   \]

The final target label, \( Y \), is thus defined by combining these two indicators using an OR operation:
\(
Y = Y_{\text{charge-off}} \lor Y_{\text{delinquency}}
\)
where \( Y = 1 \) if either charge-off or delinquency occurs within the observation window, and \( Y = 0 \) otherwise. 

If a customer does not have sufficient data within this 18-month window or lacks any credit card trades, the label is set as \texttt{None}. This approach provides a single binary outcome:
\[
Y = 
\begin{cases} 
   1 & \text{if } Y_{\text{charge-off}} = 1 \text{ or } Y_{\text{delinquency}} = 1, \\
   0 & \text{if } Y_{\text{charge-off}} = 0 \text{ and } Y_{\text{delinquency}} = 0, \\
   \text{None} & \text{if data is insufficient or unavailable}.
\end{cases}
\]
The distribution of charge-offs and delinquencies is highly imbalanced, with a much smaller proportion of customers experiencing charge-offs compared to those maintaining good credit behavior. Charge-offs occur less frequently because most customers make timely payments, while only a small percentage of customers fail to repay their debts, leading to a charge-off. This imbalance means that charge-offs are rare events in the dataset, which makes it more difficult for models to accurately predict these outcomes without specialized techniques, as the majority of cases represent customers who do not experience financial distress~\cite{alam2020investigation}.

\subsection{Ethical Consideration}
\edit{All data were fully anonymized, preserving only non-identifiable attributes and anonymized record identifiers. Each entry represents a customer’s latest credit application profile prior to \( t_0 \), with all credit attributes maintained as of the snapshot date, referred to as the run date.}

\subsection{Data Challenges}

\new{The bureau dataset is highly complex, including dense numerical codes and specialized financial terminology that pose challenges for pre-trained language models. In practice, aggregating useful features from this data has required costly manual integration or \edit{external feature purchases from the bureau itself.} The dataset functions as a domain-specific language, distinct from the natural text corpora used to train most language models. Listing~\ref{lst:data_example} illustrates the level of encoding and jargon typical of bureau records, emphasizing the need for effective domain preprocessing to make them suitable for language-based modeling.}

\begin{listing}[tb]%
\caption{Example of Bureau Credit Data}%
\label{lst:data_example}%
\begin{lstlisting}[numbers=none]
ACCT0202240P0XQ8L7J5VZK9WRN3DF4CT2Y  
BAL9988776620180715  SEG0105409000  
CAT020481 N 05062010
ENDR040017  
ACCT0302230M1R9GQ6F8TYP5ZXV2K3NBH1  
BAL7766558820170320  SEG0105409000  
CAT020481 N 06082014
ENDR050021
\end{lstlisting}
\end{listing}

\section{System Overview}

\begin{figure*}[t]  
    \centering
    \includegraphics[width=0.8\textwidth]{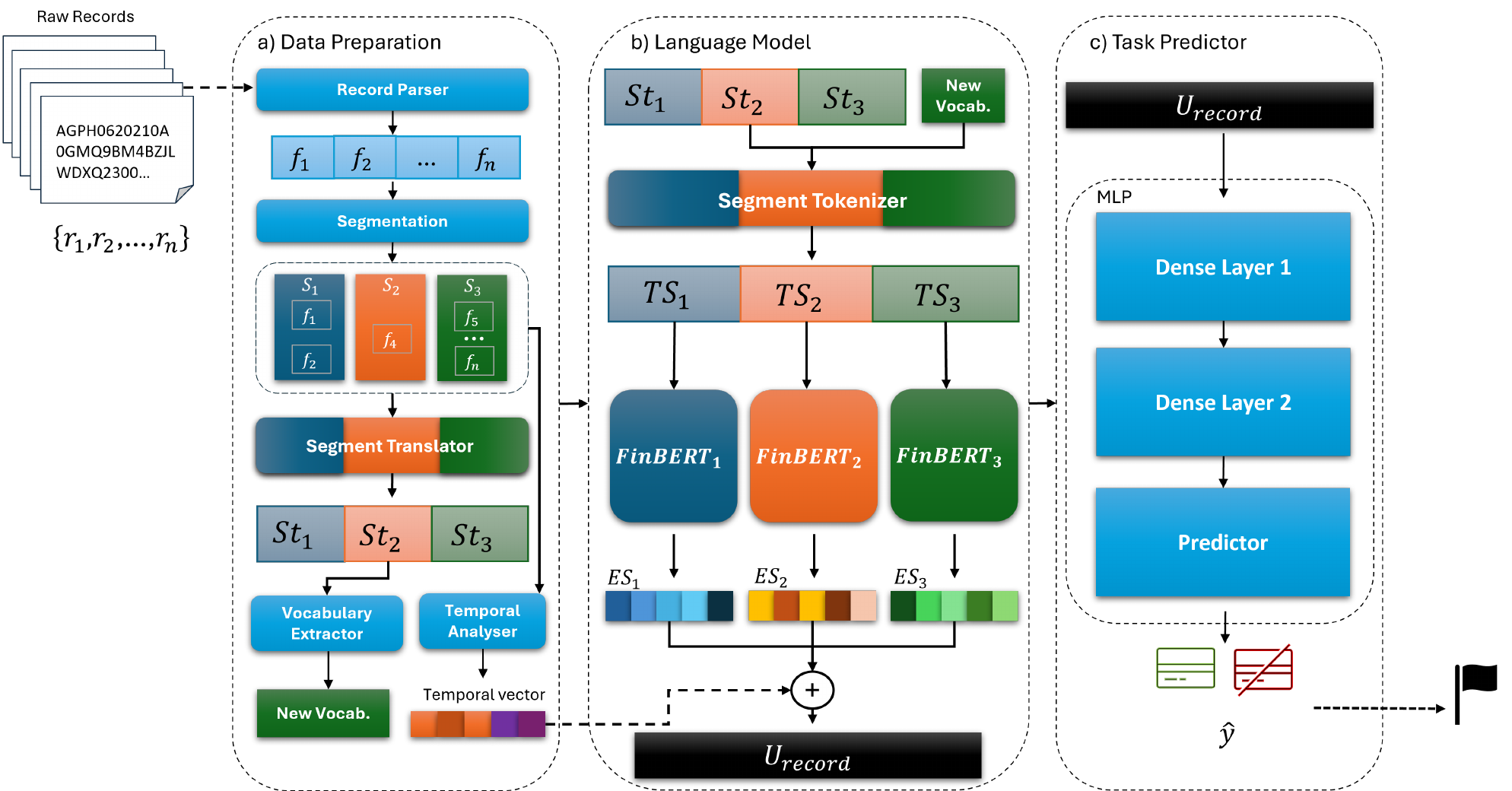}  
    \caption{\framework~System Architecture. This figure shows the three main components of \framework: (a) Data Preparation, transforming raw input into structured credit stories \( St_n \) (e.g., \( St_1, St_2, \dots \)); (b) Language Model, embedding credit stories with temporal vectors with each tokenized segment represented as \( TS_n \) (e.g., \( TS_1, TS_2, \dots \)); and (c) Task Predictor, training on these embeddings for final predictions.}
    \label{fig:system_overview}
    \vspace{-10pt}
\end{figure*}

\new{In \framework, we introduce a highly practical and cost-efficient model for credit risk assessment, designed to meet accepted performance standards in real-world applications. The system simplifies the assessment process by reducing operational overhead and improving efficiency. Its architecture comprises three key modules: (i) a data preparation stage that processes and structures raw bureau data, (ii) a language model that converts the processed information into meaningful embeddings, and (iii) a task predictor that leverages these embeddings to estimate credit risk outcomes. Each module is detailed in this section, and Figure~\ref{fig:system_overview} presents an overview of the framework.}

\subsection{Data Preparation}

This section describes the process of transforming raw bureau data into a structured format suitable for language model fine-tuning. Given the bureau data’s jargon-heavy structure and encoded language, using pre-trained language models requires careful preprocessing. These pre-trained models, typically built on standard English, cannot directly interpret the bureau’s specialized terms without modification. Research shows that fine-tuning pre-trained models is generally more efficient and effective than training from scratch in specific domains (e.g., finance) due to the substantial time and resource costs associated with model training from the ground up~\cite{li2023large, Brief2024MixingIU}. The Data Preparation Module consists of the following steps:

\subsubsection{Record Parser Module} The Record Parser Module converts unstructured bureau records into a structured vector format, where each field is positioned in a predefined order for downstream processes. For a given raw record \( R = \{r_1, r_2, \dots, r_n\} \), the parser applies a function \( f \), resulting in the structured vector:
\(
F = f(R) = \{f_1, f_2, \dots, f_n\}
\)
where \( f_i \) represents each parsed field in a fixed sequence. This transformation provides consistent ordering and schema, making subsequent segmentation and analysis feasible.

\subsubsection{Segmentation Module} The segmentation module groups related fields into segments \( S_i \), each representing different aspects of the user's behavior. For each parsed vector \( F \), a segmentation function \( g \) assigns each \( f_i \) to its respective segment:
\(
\{S_1, S_2, \dots, S_k\} = g(F)
\)
Each segment \( S_i \) aggregates fields related to specific behaviors, allowing flexibility in selecting the most relevant segments for various tasks. Key segments may be chosen for particular tasks, streamlining the analysis. In our setting, we focus on three primary segments, Trade (TR), Inquiries (IN), and Collections (CL), to capture essential elements of credit behavior. The Trade Segment offers a broad view of an individual’s credit products, showing the types and recency of obligations. The Inquiries Segment provides insight into credit application patterns, where high or recent inquiry volumes can indicate credit need or potential risk. The Collections Segment highlights past collection events, serving as a direct risk indicator. Together, these segments form the core of the individual’s credit profile, while other segments offer supplementary information.

\subsubsection{Translation Module} The translation module converts each segment \( S_i \) from encoded text into plain English, producing what we refer to as the \textit{credit story} of the user. We define a translation function \( T_i \) for each segment \( S_i \):
\(
\text{St}_i = T_i(S_i)
\)
where \( \text{St}_i \) represents the plain English equivalent of \( S_i \). The complete credit story of the user thus becomes the set \( \{\text{St}_1, \text{St}_2, \dots, \text{St}_k\} \), capturing each behavior or record in an interpretable format suitable for language model embedding.

\subsubsection{Vocabulary Extraction Module} The vocabulary extraction module identifies unique, credit-specific terms within each translated segment \( \text{St}_i \) to build a unified domain-specific vocabulary. For each segment, a vocabulary extraction function \( v_i \) extracts terms, forming individual vocabularies \( V_i \). The overall vocabulary \( V \) for tokenization is then the concatenation of these vocabularies:
\(
V = \bigcup_{i=1}^{k} V_i
\)
This expanded vocabulary enables the tokenizer to generate embeddings that incorporate credit-specific language patterns, thereby ensuring the recognition of these domain-specific terms during tokenization.

\subsubsection{Temporal Analysis Module} The temporal analysis module captures the time-dependent behavior within each segment. Each record has a base "run date" \( t_0 \), and within each segment \( S_i \), records contain fields with individual dates \( \{t_1, t_2, \dots, t_m\} \). For each date field within a record, we compute the difference between these dates and the run date \( t_0 \), creating a set of relative dates:
\(
\Delta t_j = t_j - t_0, \quad \forall t_j \in S_i
\)
For each segment, we aggregate these relative dates into a unified temporal vector \( T_{S_i} \) by calculating the minimum, maximum, and average of each date field across all records:
\(
T_{S_i} = \{\min(\Delta t), \max(\Delta t), \text{avg}(\Delta t)\}
\)
This temporal vector \( T_{S_i} \) represents each segment’s time-dependent patterns, facilitating analysis of customer behavior over time and enabling predictive insights based on temporal trends. Then, we concatenate each segment-specific temporal vector \( T_{S_i} \) to create a unified temporal behavior vector \( T_{\text{record}} \) for each record, representing the overall temporal patterns within the customer's credit history. This unified temporal vector \( T_{\text{record}} \) is expressed as:
\(
T_{\text{record}} = \{T_{S_1}, T_{S_2}, \dots, T_{S_n}\}
\)
This consolidated temporal vector \( T_{\text{record}} \) enables us to analyze and predict customer behavior over time based on the aggregated temporal trends across all segments.
Together, these modules transform the bureau’s specialized, jargon-heavy data into a format that is in standard NLP format, domain-specific, and temporally aware, ready for effective model training and interpretation.

\subsection{Language Model}
Now that we have transformed the raw bureau data into structured, plain English credit stories, we proceed to fine-tune a language model that can effectively process this financial data. Given that FinBERT is pre-trained on a large corpus of financial text, it serves as an ideal base model, as it has both an understanding of domain-specific terminology and a robust structure for analyzing financial text~\cite{yang2020finbert}. This approach minimizes the need for training from scratch and significantly enhances the model's ability to generalize on financial data, leveraging the strengths of transfer learning in this domain.

The language model module comprises three main sub-modules:

\subsubsection{Custom Tokenizer} We utilize FinBERT’s tokenizer to process each credit story \( St \). By adding domain-specific vocabulary, we ensure precise tokenization of credit-related terms. This is represented as:
    \(
    TS_i = \text{tokenize}(St_i)
    \)
    for each segment \( S_i \). This custom tokenization consolidates multi-token terms into single tokens, reducing the token count and improving the model’s comprehension of credit-specific language.

    \subsubsection{Parallel FinBERT Models} Each credit segment \( S_i \) has unique attributes, warranting separate mapping into segment-specific embeddings. For each segment, we compute the embedding:
    \(
    E_{S_i} = \text{\(FinBERT_i\)}(TS_i)
    \)
    This parallel processing maps each segment into a distinct latent space, producing embeddings \( \{E_{S_1}, E_{S_2}, \dots, E_{S_n}\} \) for all segments.

    \subsubsection{Aggregator Pooling}In this step, we combine the embeddings of all segments \( E_{S_1}, E_{S_2}, \dots, E_{S_n} \) with the temporal vector \( T_{\text{record}} \) to form a unified vector \( U_{\text{record}} \) that represents the user's credit behavior. The embeddings from each segment are concatenated along with the temporal vector to create the final vector: 
    \(
U_{\text{record}} = \left[ E_{S_1}, E_{S_2}, \dots, E_{S_n}, T_{\text{record}} \right]
\)

This unified vector \( U_{\text{record}} \) captures both the textual patterns (through segment embeddings) and temporal patterns (through the temporal vector) of the user's credit behavior. The resulting vector \( U_{\text{record}} \) is then passed to the task prediction module for further analysis.

\subsection{Task Predictor}
In the final stage of the pipeline, the aggregated representation \( U_{\text{record}} \) is used to predict the credit risk target. This unified vector, which incorporates both the segment embeddings and temporal information, is passed through a Multi-Layer Perceptron (MLP) model to make the final prediction:
\(
\hat{y} = \text{MLP}(U_{\text{record}})
\)
\edit{where \( \hat{y} \) represents the predicted probability of default.} The MLP learns to map the combined textual and temporal data to the target credit risk prediction.To address the class imbalance in the dataset, we apply weighted binary cross-entropy loss during training. This loss function assigns higher weights to the positive (default) class, which helps the model focus more on the minority class to ensure better performance in predicting defaults. 

The model is trained end-to-end for the target prediction, where the entire pipeline is jointly optimized to predict the credit risk using this weighted loss function. This ensures that the model effectively handles the class imbalance and provides accurate credit risk predictions.
\section{Experiment}
In this section, we conduct extensive experiments to evaluate the performance and practical applicability of our model. Specifically, our experiments address the following research questions:

\begin{itemize}
    \item \textbf{RQ1}: Can our language model-based framework, designed for credit risk assessment, deliver strong performance in real-world industry scenarios, meeting the ideal performance requirements for such models in practice?
    \item \textbf{RQ2}: Can the use of this model lead to cost savings in practical applications? If so, how efficient is this model in terms of cost and resource savings when compared to traditional methods in practice?
\end{itemize}

Through these experiments, we assess both the practical effectiveness and the resource efficiency of the proposed model in credit risk prediction.

\subsection{Experimental Setup}
\new{To closely replicate real-world conditions, we partitioned the data into train, validation, and two test sets: holdout and out-of-time (OOT).} The holdout set shares the same time window as the training and validation sets, while the \edit{OOT set consists of records after the training timeframe}, simulating practical deployment scenarios. Each record is anchored to an individual’s “run date,” representing the most recent date they applied for a credit card before September 2018. For training and initial validation, we selected records with a run date of February 1, 2018, or earlier, splitting them as follows:

\begin{itemize}
    \item \textbf{Train Data}: Comprising 60\% of the sampled data, this subset is used to train the model and generate baseline prediction deciles. These deciles, derived from predicted probabilities, help assess model stability by measuring the population's stability against these baselines.
    
    \item \textbf{Validation Data}: 20\% of the sampled data, used for model tuning and early stopping during training.
    
    \item \textbf{Test Data}: The remaining 20\%, reserved to evaluate overall model performance and assess generalizability. This set is split \edit{into OOT (after training timeframe) and Holdout (same as training timeframe) sets.} 
\end{itemize}

\textbf{Data Imbalance}: The data is highly imbalanced, with a target event rate of less than 5\%. A slightly lower event rate is expected in the OOT set, reflecting the 2018 market conditions, which saw fewer overall defaults compared to previous years.

\edit{The industry standard for assessing model performance is the bureau-provided credit score, calculated using an extensive set of features extracted from credit records. This credit score has an AUC of approximately 0.8 for default prediction on our particular dataset,} serving as the benchmark for practical credit risk assessment. \new{We introduce a new baseline and direction for credit risk modeling, demonstrating the effectiveness of language-based approaches for understanding bureau data. This work paves the way toward developing a financial foundation model capable of supporting broader decision-making in credit and risk management.}

\subsection{Results}
\subsubsection{Acceptable performance of industrial language model in credit default prediction(RQ1).} We implemented an iterative model construction framework to push the boundaries of our language model, aiming for performance competitive with industry baselines, while avoiding the costly steps typical of traditional models. Our model underwent six key versions, as shown in Figure~\ref{fig:model-iterations}:

\begin{figure}[h]
    \centering
    \includegraphics[width=0.39\textwidth]{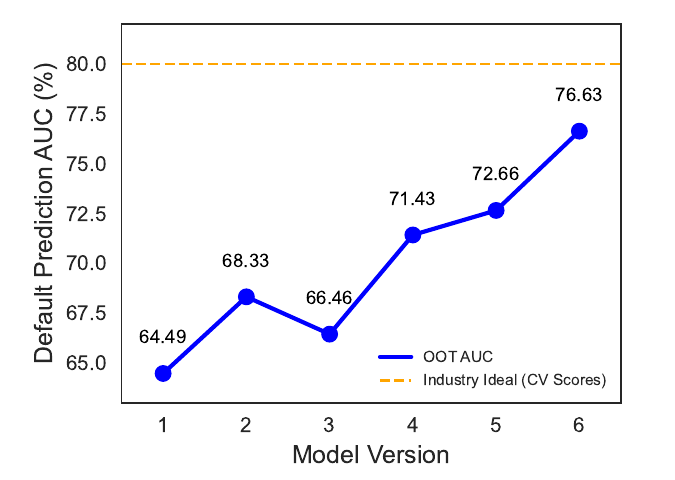} 
    \caption{\new{Performance Trend of Model Versions. This chart shows the steady improvement in AUC scores across model development stages, with our final version achieving close alignment to the industry's ideal baseline, reflecting the potential of \framework~in credit risk prediction.}}
    \label{fig:model-iterations}
    \vspace{-12pt}
\end{figure}

\begin{itemize}
    \item \textbf{Version 1}: The initial model pipeline was established without a custom tokenizer. Each segment had a separate classifier, and the final prediction was determined by majority voting. Due to resource limitations, training was conducted incrementally across smaller data buckets.
    
    \item \textbf{Version 2}: A custom tokenizer was added to incorporate specific credit vocabulary, improving the relevance of the tokenized data.
    
    \item \textbf{Version 3}: We introduced label balancing by oversampling positive cases to address the label imbalance, achieving a 70-30 balance. This adjustment, however, showed reduced effectiveness and was modified in later Versions.
    
    \item \textbf{Version 4}: The tokenization was optimized to remove redundancies, and the training loop was modified to use a single, comprehensive dataset, making it more comparable to traditional models. Label balancing was adjusted to 90-10, closely reflecting the real-world distribution.
    
    \item \textbf{Version 5}: The pooling mechanism was shifted to aggregate embeddings directly, feeding a unified vector into an expanded MLP layer. This change improved the capacity to handle the aggregated inputs and removed the label balancing for a more natural distribution.
    
    \item \textbf{Version 6}: Our champion model, depicted in Figure~\ref{fig:system_overview}, included temporal information injected into the text embeddings, further enhancing predictive accuracy.
\end{itemize}

\begin{figure}[h]
    \centering
    \includegraphics[width=0.39\textwidth]{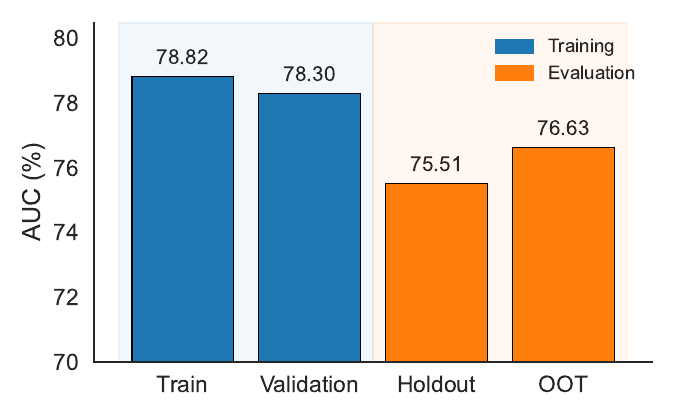} 
    \caption{\new{AUC of Champion Model Across Data Splits}}
    \label{fig:auc_metrics}
    \vspace{-10pt}
\end{figure}

After these six Versions, our final model stands at just a 3.63\% AUC gap behind the industrial baseline, showcasing the promise of language models in credit risk prediction. This automated, scalable approach demonstrates the potential for generalization across other tasks within the credit domain. Figure~\ref{fig:auc_metrics} shows the performance of our final version model over different data splits.

\subsubsection{Cost Efficiency of Language Models in Credit Default Prediction (RQ2).}

Our industrial-scale evaluation shows that \framework~offers notable cost efficiency compared to traditional approaches in credit risk modeling. Due to the confidentiality of the framework and its commercial deployment, exact numerical values cannot be disclosed. Instead, we highlight key domains where \framework~demonstrates measurable efficiency and potential for further optimization.  

\begin{itemize}
    \item \textbf{Data Licensing Efficiency:} Traditional models rely on bureau-aggregated features whose cost scales with data volume and purchased bundles. \framework~ operates directly on raw bureau data, removing repeated licensing expenses and maintaining near-linear scalability.

    \item \textbf{Multi-Task Learning Scalability:} Conventional systems train separate models for each task, increasing maintenance cost. \framework~serves as a unified foundation model supporting multiple credit decision tasks within one framework, improving scalability and resource use.

    \item \textbf{Training and Compute Efficiency:} While LLMs incur higher initial training cost, this is amortized across many tasks. Fine-tuning and inference require far fewer resources than retraining several independent models.

    \item \textbf{Feature Engineering Automation:} Traditional workflows depend on feature engineering steps, adding latency and complexity. \framework~automates this step by extracting representations from unstructured credit narratives, reducing pre-processing overhead.

    \item \textbf{Cost Saving and Practical Impact:} The framework scales linearly with the cost of data acquisition, purchased features, and the number of customers. It can, in theory, extract unlimited features directly from text without licensing. In large-scale markets such as Canada, this enables potential savings in the order of millions of dollars annually through unified modeling and reduced data dependency.
\end{itemize}

These observations indicate that \framework~provides a promising foundation for scalable and cost-efficient credit modeling, with opportunities for further enhancement through model scaling and process optimization.

\section{Conclusion}

\new{
Our study presents language models as automated systems for end-to-end credit risk assessment, capable of performing end-to-end evaluation directly from bureau data. With \framework, we establish a new baseline for automated credit modeling, showing that such models can interpret financial patterns and support reliable decision-making without manual intervention. \edit{This direction can be further advanced to outperform bureau credit score baselines through larger-scale training and extend the architecture to capture richer financial signals.} As bureau data underpin many key decisions in banking, this framework can naturally extend to those tasks. Looking ahead, this work lays the groundwork for developing multi-task and multi-domain financial foundation models that unify risk assessment and decision support across the broader financial ecosystem.
}

%

\bibliography{aaai2026}

\appendix

\end{document}